\documentclass[11pt,a4paper]{article}
\usepackage[hyperref]{naacl2018}
\usepackage{times}
\usepackage{latexsym}
\usepackage{url}
\usepackage{amsmath}
\usepackage{multirow}
\usepackage{url}
\usepackage{graphicx}
 \usepackage{subfig}

\aclfinalcopy 

\title{Interpreting Verbal Irony: Linguistic Strategies and the Connection to the Type of Semantic Incongruity}

\author{Debanjan Ghosh Elena Musi Kartikeya Upasani Smaranda Muresan}

\author{
  Debanjan Ghosh\thanks{Part of the research was carried out while Debanjan was a Ph.D. candidate at Rutgers University.}\textsuperscript{1}, 
  Elena Musi\textsuperscript{2}, 
  Kartikeya Upasani\textsuperscript{3},
  Smaranda Muresan\textsuperscript{4} \\
  \textsuperscript{1} McGovern Institute for Brain Research, MIT, Cambridge, MA  \\
  \textsuperscript{2} University of Liverpool, Liverpool, UK\\
    \textsuperscript{3} Facebook Conversational AI, CA\\
    \textsuperscript{4} Data Science Institute, Columbia University, New York, NY\\
    \tt{dg513@mit.edu, elena.musi@liverpool.ac.uk,}\\
    \tt{kart@fb.com, smara@columbia.edu}
    } 

\date{}
\begin{document}
\maketitle
\begin{abstract}
Human communication often involves the use of verbal irony or sarcasm, where the speakers usually mean the opposite of what they say. To better understand how verbal irony is expressed by the speaker and interpreted by the hearer we conduct a crowdsourcing task: given an utterance expressing verbal irony, users are asked to verbalize their interpretation of the speaker's ironic message. We propose a typology of linguistic strategies for verbal irony interpretation and link it to various theoretical linguistic frameworks. We design computational models to capture these strategies and present empirical studies aimed to answer three questions: (1) what is the distribution of linguistic strategies used by hearers to interpret ironic messages?; (2) do hearers adopt similar strategies for interpreting the speaker's ironic intent?; 
and 
 (3) does the type of semantic incongruity in the ironic message (explicit vs. implicit) influence the choice of interpretation strategies by the hearers? 
\end{abstract}

\section{Introduction}

It is well understood that recognizing whether a speaker is ironic or sarcastic is essential to understanding their actual sentiments and beliefs. For instance, the utterance ``pictures of holding  animal carcasses are so \textit{flattering}" is an expression of verbal irony, where the speaker has a negative sentiment towards ``pictures of holding animal carcasses", but uses the positive sentiment word ``flattering". This inherent characteristic of verbal irony is called semantic incongruity --- incongruity between the literal evaluation and the context (e.g., between the positive sentiment words and the negative situation in this example). Most NLP research on verbal irony or sarcasm has focused on the task of \emph{sarcasm detection} treating it as a binary classification task using either the utterance in isolation or adding contextual information such as conversation context, author context, visual context, or cognitive features \cite{davidov2010,maynard2014cares,wallace2014humans,joshi2015harnessing,bamman2015contextualized,muresanjasist2016,amir2016modelling,mishra-etal-2016-harnessing,ghoshmagnet2017,felbo2017,ghosh2017role,hazarika2018,tay2018,ghosh2018sarcasm,oprea2019exploring}. 
Such approaches have focused their analysis on the speakers' beliefs and intentions for using irony \cite{attardo2000ironya}. 
However, sarcasm and verbal irony are types of interactional phenomena with specific perlocutionary effects on the hearer \cite{haverkate1990speech}. 
Thus, we argue that, besides recognizing the speaker's sarcastic/ironic intent, it is equally important to understand \emph{how the hearer interprets} the speaker's sarcastic/ironic message. For the above utterance, the strength of negative sentiment perceived by the hearer depends on whether they interpret the speaker's actual meaning as ``picture \dots are \textbf{not flattering}'' vs. ``pictures \dots are \textbf{so gross}'' (Table \ref{table:sarcesboth}).
The intensity of negative sentiment is higher in the latter interpretation than in the former.
\citet{kreuz2000production} noted that most studies in linguistics and psychology have conducted experiments analyzing reaction times \cite{gibbs1986psycholinguistics,katz2004saying} or situational context \cite{ivanko2003context}, featuring a setup with \emph{in vitro} data aimed at testing the validity of specific theories of irony.
Instead, our study adopts a \emph{naturalistic} approach to understand hearers' reception of irony looking at what linguistic strategies are recurrently used by hearers to interpret the non-literal meaning underlying ironic utterances.  

We leverage the crowdsourcing task introduced by \newcite{ghoshguomuresan2015} for their work on detecting whether a word has a literal or sarcastic interpretation, later adopted by \newcite{peled2017sarcasm}. The task is framed as follows: given a speaker's ironic message, five annotators (e.g., Turkers on Amazon Mechanical Turk (MTurk)) are asked to verbalize their interpretation of the speaker's ironic message (i.e., their understanding of the speaker's intended meaning) 
(see Table \ref{table:sarcesboth}; S$_{im}$ denotes the speaker's ironic message, while H$_{int}$ denotes the hearer's interpretation of that ironic message). The crowdsourcing experiments are reported in Section~\ref{section:data}.

\begin{table*}[ht] 
\centering
\begin{small}
\begin{tabular}{p{3.5cm}p{3.5cm}p{3.5cm}p{3.5cm}}
\hline
 S$_{im}$ & H$^{1}_{int}$ & H$^{2}_{int}$ & H$^{3}_{int}$  \\
\hline
1. Ed Davey is such a passionate, inspiring speaker & Ed Davey is a boring, uninspiring speaker & Ed Davey is such a dull, monotonous speaker & Ed Davey is not a passionate, inspiring speaker \\
\hline
2. can't believe how much captain America looks like me  & I wish I looked like Captain America. I need to lose weights & can't believe how much captain America looks different from me & I don't, but I wish I looked like Captain America \\ 
\hline
3. Pictures of you holding dead animal carcasses are so flattering & Hate hunting season and the pictures of you holding dead animal are so gross  &  Pictures of you holding dead animal carcasses is an unflattering look & Pictures of you holding dead animal carcasses are not flattering \\
\hline
\end{tabular}
\end{small}
\caption{\small{Examples of speaker's ironic messages (S$_{im}$) and interpretations given by 3 Turkers (H$^{i}_{int}$).}}	
\centering
\label{table:sarcesboth}
\end{table*}

The paper makes three contributions. 
First, we propose a data-driven \emph{typology of linguistic strategies} that hearers use to interpret ironic messages and discuss its relevance in verifying theoretical frameworks of irony (Section \ref{section:strategies}). 
Second, we propose computational models to capture these strategies (Section \ref{section:empirical}). 
Third, we present two studies that aim to answer two questions: (1) does the type of semantic incongruity in the ironic message (explicit vs. implicit; see Section \ref{section:ExpIm}) influence the choice of interpretation strategies by the hearers? (Section \ref{section:tweets}); (2) do interpretation strategies of verbal irony vary by hearers?
We make all datasets and code available.\footnote{https://github.com/debanjanghosh/interpreting\_verbal\_irony}

\section{Datasets of Speakers' Ironic Messages and Hearers' Interpretations}  \label{section:data}



To generate a parallel dataset of speakers' ironic messages and hearers' interpretations we conduct a crowdsourcing experiment. Given a speaker's ironic message (S$_{im}$), five Turkers (hearers) on MTurk are asked to verbalize their interpretation of the speaker's ironic message (i.e., their understanding of the speaker's intended meaning) (H$_{int}$). 
The design of the MTurk task was first introduced by \newcite{ghoshguomuresan2015}, who use the resulting dataset to identify words that can have both a literal and a sarcastic sense. \newcite{peled2017sarcasm} employed similar design to generate a parallel dataset to use for generating interpretations of sarcastic messages using machine translation approaches. They use  skilled annotators in comedy writing and literature paraphrasing and give them the option not to rephrase (we refer to  \newcite{peled2017sarcasm}'s dataset as $SIGN$). We perform this new crowdsourcing task and do not rely entirely on the above two datasets for two reasons: (1) we focus on verbal irony, and (2) we always require an interpretation from the Turkers. 
Unlike the above two studies, the main goal of our research is to analyze the linguistics strategies employed by hearers in interpreting verbal irony.



We collected messages that express verbal irony from Twitter using the hashtags \#irony, \#sarcastic,  and \#sarcasm. 
We chose Twitter as a source since the presence of the hashtags allows us to select sentences where the speaker's intention was to be ironic. Furthermore, even though Twitter users cannot be considered representative of the entire population, they are unlikely to be skewed with respect to topics or gender. 
We manually checked and 
kept 1,000 tweets that express verbal irony.
We do not draw any theoretical distinction between sarcasm and irony since we cannot assume that Twitter users also differentiate between \#irony and \#sarcasm, blurred even in scholarly literature. The Turkers were provided with detailed instructions and examples of the task including the standard definition of verbal irony taken from the Merriam-Webster dictionary (``use of words to express something other than and especially the opposite of the literal meaning''). We decided to suggest them a guiding definition for two reasons. First, hearers do not usually focus on literal vs. non literal meaning, as shown by studies measuring processing times for both types of statements \cite{inhoff1984contextual}. Therefore, when asked to rephrase the speakers' intended meaning, hearers would have probably come up with sentences expressing the speaker's imagined discursive goals, rather than disclosing their perceived literal meaning. Second, it is reasonable to assume that Turkers would have looked up the standard meaning of ironic utterance given by an online dictionary to ease up their task, possibly coming up with biased definitions. 

The Turkers were instructed to consider the entire message in their verbalization to avoid asymmetry in length between the S$_{im}$ and  H$_{int}$. 
We obtained a dataset of 5,000  S$_{im}$-H$_{int}$ pairs where five Turkers rephrase each S$_{im}$. A total of 184 Turkers participated in the rephrasing task. Table \ref{table:sarcesboth} shows examples of speaker's ironic messages (S$_{im}$) and their corresponding hearers' interpretations (H$^{i}_{int}$). 
Next, we ran a second MTurk task to verify whether the generated H$_{int}$ messages are plausible interpretations of the ironic messages. This time we employ three Turkers per task and only Turkers who were not involved in the content generation task were allowed to perform this task. We observe that Turkers labeled 5\% (i.e., 238 verbalizations) of H$_{int}$s as invalid and low quality (e.g., wrong interpretation). For both tasks, we allowed only qualified Turkers (i.e., at least 95\% approval rate and 5,000 approved HITs), paid 7 cents/task and gave sixty minutes to complete each task. The final dataset contains 4,762 pairs S$_{im}$-H$_{int}$. 
\section{Semantic Incongruity in Ironic Messages: Explicit vs. Implicit} \label {section:ExpIm}
 \newcite{attardo2000ironya} and later \newcite{burgers2010verbal} distinguish between two theoretical aspects of irony: \textit{irony markers} and \textit{irony factors}. Irony markers are meta-communicative signals, such as interjections or emoticons that alert the reader that an utterance might be ironic.
In contrast, irony factors cannot be removed without destroying the irony, such as the incongruity between the literal evaluation and its context (``semantic incongruity'').
 Incongruity expresses the contrast between the conveyed \textit{sentiment} (usually, positive) and the targeted \textit{situation} (usually, negative). This contrast can be explicitly or implicitly expressed in the ironic message. 
Following \citet{karoui2017exploring}, we consider that semantic incongruity is explicit, when it is lexicalized in the utterance itself (e.g., both the positive sentiment word(s) and the negative situation are available to the reader explicitly). On Twitter, beside sentiment words, users often make use of hashtags (e.g., ``Studying 5 subjects \dots \#worstsaturdaynight'') or an image (e.g., ``Encouraging how Police feel they're above the law. URL''; the URL shows a police car not paying parking) to express their sentiment. We consider these cases as explicit, since the incongruity is present in the utterance even if via hashtags or other media.  For implicit incongruity, we consider cases where one of the two incongruent terms (``propositions" in \citet{karoui2017exploring}) is not lexicalized and has to be reconstructed from the context (either outside word knowledge or a larger conversational context). For example ``You are such a nice friend!!!'', or ``Driving in Detroit is fun ;)" are cases of ironic messages where the semantic incongruity is implicit. 
Based on these definitions of explicit and implicit incongruity,  two expert annotators annotated the S$_{im}$-H$_{int}$ dataset (1000 ironic messages) as containing explicit or implicit semantic incongruity. The inter-annotator agreement was $\kappa$=0.7, which denotes good agreement similar to \newcite{karoui2017exploring}. The annotation showed that 38.7\%  of the ironic messages are explicit, while 61.3\% are implicit.  In the following section we propose a typology of linguistic strategies used in hearers' interpretations of speakers' ironic messages and in section \ref{section:propstrategies} we discuss the correlation of linguistic strategies with the type of semantic incongruity.


\section{Interpreting Verbal Irony: A Typology of Linguistic Strategies} \label{section:strategies}



Given the definition of verbal irony, we would expect that Turkers' interpretation of speaker's ironic message will contain some degree of opposite meaning with respect to what the speaker has said. However, it is unclear what linguistic strategies the Turkers will use to express that. 
To build our typology, from the total set of S$_{im}$-H$_{int}$ pairs obtained through crowdsourcing (i.e., 4,762 pairs; see Section \ref{section:data}) we selected a $dev$ set of 500 S$_{im}$-H$_{int}$ pairs. 
Our approach does not assume any specific theory or irony, but it is data-driven: a linguist expert in semantics and pragmatics analyzed the $dev$ set to formulate the lexical and pragmatic phenomena attested in the data. The assembled typology is, thus, the result of a bottom-up procedure. A S$_{im}$-H$_{int}$ pair can be annotated with more than one strategy. The core linguistic strategies are explained below and synthesized in Table \ref{table:distrigold}.


%

\subsection{Linguistic Strategies}
\paragraph{Lexical and phrasal antonyms:} This category contains lexical antonyms (e.g., ``love''  $\leftrightarrow$ ``hate'', ``great'' $\leftrightarrow$ ``terrible'') as well as indirect antonyms \cite{fellbaum1998wordnet}, where the opposite meaning can only be interpreted in context (e.g., ``{passionate} speaker'' $\rightarrow$ ``{boring} speaker''; Table \ref{table:sarcesboth}). Although the typical antonym of  ``passionate'' is ``unpassionate'', ``boring'' works in this context as a lexical \emph{opposite} since a speaker who is passionate entails that he is not boring. 
Besides lexical antonyms, Turkers sometimes use
antonym phrases (e.g., ``I can't wait'' $\rightarrow$ ``not looking forward'', ``I like (to visit ER)'' $\rightarrow$ ``I am upset (to visit ER) '').


\paragraph{Negation:} Here, Turkers negate the main predicate. This strategy is used in the presence of copulative constructions where the predicative expression is an adjective/noun expressing sentiment (e.g., ``is great'' $\rightarrow$ ``is \textbf{not} great'') and of verbs expressing sentiment (e.g., ``love'' $\rightarrow$ ``\textbf{do not} love'') or propositional attitudes (e.g., ``I wonder'' $\rightarrow$ ``I \textbf{don't} wonder''). 
\begin{table}
\begin{center}
\begin{small}
\begin{tabular}{lc}
\hline
Typology & Distribution (\%)\\
\hline
\textbf{Antonyms} & \\
 - lexical antonyms &  (42.2) \\
 - antonym phrases &  (6.0)\\
\textbf{Negation} & \\
 - simple negation &  (28.4) \\
\textbf{Antonyms OR Negation} & \\
 - weakening sentiment &  (23.2)\\
 - interrogative $\rightarrow$ declarative  &  (5.2)\\
 - desiderative constructions &  (2.8)\\
\textbf{Pragmatic inference}&  (3.2)\\
\hline
\end{tabular}
\end{small}
\end{center}
\caption{Typology of linguistic strategies and their distribution (in \%) over the $dev$ set}
\label{table:distrigold}
\end{table}
\paragraph{Weakening the intensity of sentiment:} The use of negation and antonyms is sometimes 
accompanied by two strategies that reflect a weakening of sentiment intensity. First, when S$_{im}$ contains words expressing a high degree of positive sentiment, the hearer's interpretation replaces them with more neutral ones (e.g., ``I \textbf{love} it'' $\rightarrow$ ``I don't \textbf{like} it''). Second, when S$_{im}$ contains an intensifier, it is eliminated in the Turkers' interpretation. Intensifiers  specify the degree of value/quality expressed by the words they modify \cite{mendez2008special} (e.g., ``cake for breakfast. \textbf{so} healthy'' $\rightarrow$ ``cake for breakfast. \textbf{not} healthy''). 
 



\paragraph{Interrogative to Declarative Transformation (+ Antonym/Negation):} This strategy, used mostly in conjunction with the negation or antonym strategies, consists in replacing the interrogative form with a declarative form, when S$_{im}$ is a rhetorical question (for brevity, $RQ$) (e.g., ``don't you \textbf{love} fighting?'' $\rightarrow$ ``I \textbf{hate} fighting''). 

 

\paragraph{Counterfactual Desiderative Constructions:} When the ironic utterance expresses a positive/negative sentiment towards a past event (e.g., ``glad you relayed this news'') or an expressive speech act (e.g., ``thanks $X$ that picture needed more copy'') the hearer's interpretation of intended meaning is expressed through the counterfactual desiderative constructions \textit{I wish (that) p} (``\textbf{I wish} you hadn't relayed \dots'', ``\textbf{I wish} $X$ didn't copy \dots'').
Differently from antonymic phrases, this strategy stresses on the failure of the speaker's expectation more than on their commitment to the opposite meaning.

\paragraph{Pragmatic Inference:} In addition to the above strategies, there are cases where the interpretation calls for an inferential process to be recognized. For instance, ``made 174 this month \dots \textit{I'm gonna buy a yacht!}'' $\rightarrow$ ``made 174 this month \dots \textbf{I am so poor}". 
%
The distribution of the strategies on the $dev$ set is represented in Table \ref{table:distrigold}.


 \subsection{Links to Theoretical Frameworks} \label{section:theory}

In linguistic literature many different approaches to irony have been provided. Here we focus on the three accounts (w.r.t. examples from S$_{im}$-H$_{int}$ corpus) that bear a different views on pragmatic factors. According to \citet{grice1975syntax}, ironic messages are uttered to convey a meaning opposite to that literally expressed, flouting the conversational maxim of quality ``do not say what you believe to be false''. In verbal irony, the violation of the maxim is frequently signaled by ``the opposite'' of what is said literally (e.g., intended meaning of ``carcasses are flattering'' is they are gross; Table \ref{table:sarcesboth}).  
The linguistic strategies of \textit{antonyms} (e.g. ``worst day of my life") and simple \textit{negation} (``yeap we totally dont drink alcohol every single day''[...]) cover the majority of the S$_{im}$-H$_{int}$ corpus and seem to fit the Gricean \cite{grice1975syntax} account of irony, since the hearer seems to have primarily recognized the presence of semantic incongruity. 
However, as touched upon by \newcite{giora1995irony}, \textit{antonyms} and \textit{direct negation} are not always semantically equivalent strategies, since the second sometimes allows a graded interpretation: if ``x is not encouraging", it is not necessarily bad, but simply ``x $<$ encouraging''. Such an implicature is available exclusively with items allowing mediated contraries, such as sentiment words \cite{horn1989natural}. Direct negation with sentiment words implies that just one value in a set is negated, while the others are potentially affirmed.
The spectrum of interpretations allowed by negation as a rephrasing strategy indicates that hearers recognize that the \emph{relevance} of the ironic utterance in itself plays a role next to what the utterances refers to (if the rephrased utterance is intended as ``x is not encouraging at all'', the perceived irrelevance of the corresponding ironic  utterance is more prominent than in ``x is not very encouraging"). 
The fact that the interpretation of irony has a propositional scope is even clearer when the ironic sentence in interrogative form (``and they all lived happily ever after ?") is rephrased as a declarative (e.g. ``I doubt they all lived happily ever after"): the hearers recognizes that the question has a rhetoric value since otherwise contextually irrelevant. 
The intentional falsehood of Gricean analysis is also not deemed by \newcite{sperber1986relevance,wilson2012explaining} as a necessary and sufficient condition for irony. According to their theory of \emph{echoic mentioning}, irony presupposes the mention to the inappropriateness of the entire sentence: in asserting ``awesome weather in Scotland today" the speaker does not simply want to express that the weather was horrible but he signals that assuming that the weather would be nice was irrelevant and, thus, ridiculous. \citet{kreuz1989sarcastic} expand the Relevance Theory approach talking about \emph{echoic reminding} to account for cases such as ``could you be just a little louder, please? My baby isn't trying to sleep" where the extreme politeness reminds the hearer that the question is indeed a request and that the mother bears a certain stance and has certain expectations towards the addressee.
Similarly, the use of the \emph{pragmatic inference} strategy cannot be fully explained in Gricean terms: the rephrase ``made 174 this month \dots I am so poor'' for ``made 174 this month \dots I am gonna buy a yatch'' more than pointing to the presence of lexical incongruity, show that the hearers knows for background knowledge that the assertion of ``buying a yatch'' is completely irrelevant in the context of a low salary situation. 
Rephrasing strategies using counterfactual desiderative constructions (e.g. ``I really wish my friends and family would check up on my after yesterday's near death experience") show, instead, that the interpretation of irony involves an \emph{echoic reminding} to the speaker's (social) expectations which failed to be fulfilled.
Overall, using the results of our crowdsourcing experiment with main existing theories of irony, it turns out that the theories have a complementary explanatory power. In Section \ref{section:tweets} we investigate weather this situation might relate to the presence of explicit/implicit irony.

\section{Empirical Analysis of Interpretation Strategies} \label{section:empirical}

Here our goal is to perform a comparative empirical analysis to understand how hearers interpret verbal irony. To accomplish this, we propose computational models to automatically detect these linguistic strategies in two datasets: (1) S$_{im}$ -H$_{int}$ dataset and (2) the $SIGN$ dataset. As stated in Section \ref{section:data}, albeit for a different purpose, the task designed in \newcite{peled2017sarcasm} is identical to ours: they used a set of 3,000 sarcastic tweets and collected five interpretation verbalization, including an option to just copy the original message if it was not deemed ironic.  They used workers skilled in comedy writing and literature paraphrasing. $SIGN$ contains 14,970 pairs. To evaluate our models, we asked two annotators to annotate two $test$ sets of 500 pairs each from the S$_{im}$ -H$_{int}$ and the $SIGN$ dataset (i.e., denoted by $SIGN_{test}$), respectively. Note, the $test$ set for the S$_{im}$ -H$_{int}$ has no overlap with the $dev$ set of 500 S$_{im}$-H$_{int}$ pairs used to identify the strategies (Section \ref{section:strategies}). Agreement between the annotators for both sets is high with $\kappa>$ 0.9.  In $SIGN_{test}$, 79 instances were just copies of the original message, which we eliminated, thus the $SIGN_{test}$ contains only 421 instances. 

%

\subsection{Computational Methods} \label{subsection:stratcompute}

\begin{table*}[h]
\centering
\begin{small}
\begin{tabular}{p{3 cm}p{.6cm}p{.6cm}p{.6cm}p{.6cm}p{.6cm}p{.6cm}p{.6cm}p{.6cm}p{.6cm}}
\hline
 & \multicolumn{3}{c} {$dev$} & \multicolumn{3}{c} {$test$} & \multicolumn{3}{c} {$SIGN_{test}$} \\
Strategies & P & R & F1 & P & R & F1 & P & R & F1 \\
\hline
Lex\_ant  
&   89.0	& 95.7 &	92.2 & 97.2 & 89.9 & 93.4 & 89.4 & 97.9 &	93.5\\
Simple\_neg 
&  92.0 &	89.4 &	90.7 & 	88.3 & 88.3 & 88.3 & 93.3 &	91.2 &	92.2\\
AN\_weaksent 
& 93.6 &	87.9 & 90.7 &   95.0	& 91.9 &	93.4 & 93.3 & 87.5 &	90.3	\\
AN$_{I \rightarrow D}$ 
&  53.1 & 65.4 & 58.6 & 80.0 & 0.44 & 57.2  & 85.7 & 70.6 & 77.4    \\
AN\_desiderative 
& 100.0	 & 92.9  & 	96.3   & 100.0 & 100.0 & 100.0 & 100.0 & 66.7 & 80.0\\
AntPhrase+PragInf 
& 86.2 & 53.2 & 65.8 & 70.7 & 	85.3 &	77.4 & 89.5	& 68.0	& 77.3 \\
\hline
\end{tabular}
\caption{Evaluation of Computational Methods on $dev$, $test$ and ${SIGN_{test}}$ set (in \%)}
\label{table:devprf1}
\end{small}
\end{table*}

\paragraph{Lexical Antonyms.} To detect whether an S$_{im}$-H$_{int}$ pair uses the \emph{lexical antonyms} strategy, we first need to build a resource of lexical antonyms. 
We use the MPQA sentiment Lexicon \cite{wilson2005recognizing}, \newcite{hu2004}'s opinion lexicon, antonym pairs from \newcite{Mohammad13}, antonyms from WordNet, and pairs of opposite verbs from Verbocean \cite{chklovski2004verbocean}. 

Given this lexicon of lexical antonyms, the task is now to detect whether a given S$_{im}$-H$_{int}$ pair uses the \emph{lexical antonyms} strategy. We use a heuristic approach based on word-alignment and dependency parsing (similar to contradiction detection \cite{de2008finding}). Word-to-word alignments between S$_{im}$-H$_{int}$ are extracted using a statistical machine translation (SMT) alignment method - IBM Model 4 with HMM alignment from Giza++ \cite{och}. We consider a lexical antonym strategy if: 1) antonym words are aligned; 2) they are the roots of the respective dependency trees or if the nodes modified by the lexical antonyms are the same in their respective trees (e.g., `can you show any \textbf{more} of  steelers'' $\rightarrow$ ``show \textbf{less} of  steelers'', the candidate lexical antonyms are \textit{more} and \textit{less} and they are the objects of the same predicate in S$_{im}$-H$_{int}$: \textbf{show}). Out of 211 S$_{im}$-H$_{int}$ pairs that are marked as having \emph{lexical antonym} strategy ($dev$ set), 12 instances are identified by only the dependency parses, 67 instances by the word-alignments, and 100 instances by both (P/R/F1 scores  are 92.1\%, 77.7\% and 84.3\%), respectively on $dev$ dataset.
However, sometimes both dependency and word-alignment methods fail. In ``circling down the bowl. \textbf{Yay}'' $\rightarrow$ ``circling down the bowl.  \textbf{awful}'', although the lexical antonyms  \textbf{yay} and \textbf{awful} exist, neither the alignment nor the dependency trees can detect it (25 such instances in the $dev$ set). To account for this, after having run the dependency and alignment methods, we also just search whether a S$_{im}$-H$_{int}$ pair contains a lexical antonym pair. This improves the final recall and on the $dev$ set we achieve 89.0\% precision, 95.7\% recall, and 92.2\% F1 on $dev$ dataset (Lex\_{ant} Strategy; Table \ref{table:devprf1} show results both on $dev$ and the  $test$ sets). Note, just searching whether a lexical antonym pair is present in a S$_{im}$-H$_{int}$ pair results in low precision (58.6\%) but high recall (80\%).

\paragraph{Simple negation.} This strategy (denoted as Simple\_neg in Table \ref{table:devprf1} and Table \ref{table:distribution}) involves identifying the presence of negation and its scope. Here, however, the scope of negation is constrained since generally Turkers negated only a single word (i.e., ``love'' $\rightarrow$ ``\textbf{not} love''). Thus our problem is easier than the general problem of finding the scope of negation \cite{li2018learning,qian2016speculation,fancellu2016neural}.
We use 30 negation markers from \newcite{reitan2015negation} to find negation scope in tweets. We first detect whether a negation marker appears in either H$_{int}$ or S$_{im}$, but not in both (negation can appear in S$_{im}$ for ironic blame) 
If the marker is used, we extract its parent node from the dependency tree, and if this node is also present in the other utterance, then \emph{Negation} strategy is selected. For instance, in ``\textbf{looks} just like me'' $\rightarrow$ ``does \textbf{not} \textbf{look} like me'', the negation \textbf{not} is modifying the main predicate \textbf{looks} in H$_{int}$, which is also the main predicate in  S$_{im}$ (words are lemmatized). In the next section, we discuss if the parent nodes are not the same but similar and with different sentiment strength.

\paragraph{Weakening the intensity of sentiment.} The first strategy --- replacing words expressing a high degree of positive/negative sentiment with more neutral ones (`I \textbf{love} being sick'' $\rightarrow$ ``I \textbf{don't} \textbf{like} being sick)---, is applied only in conjunction with the negation strategy.
We measure the difference in strength using the Dictionary of Affect \cite{whissell1986dictionary}. Out of 31 S$_{im}$-H$_{int}$ pairs in the $dev$ set, we automatically identify 28 interpretations that use this approach.  
For the second strategy --- removing the intensifier (I am \textbf{really} happy'' $\rightarrow$ ``I am disappointed') ---,  we first determine whether the intensifier exists in S$_{im}$ and is eliminated from H$_{int}$. We use only adjective and adverb intensifiers from \newcite{taboada2011lexicon}, primarily to discard conjunctions  such as ``so'' (``no water \textbf{so} I can't wash \dots''). 
This strategy is used together with both \emph{lexical antonyms} and \emph{Simple negation} strategies. 
For a candidate S$_{im}$-H$_{int}$ pair, if the \emph{lexical antonym} strategy is selected and $a_{S}$ and $a_{H}$ are the lexical antonyms, we determine whether any intensifier modifies $a_{S}$ and no intensifier modifies $a_{H}$. 
If the \emph{Negation} strategy is selected, we identify the negated term in the H$_{int}$ and then search its aligned node from the S$_{im}$ using the word-word alignment. Next, we search in the S$_{im}$ if any intensifier is intensifying the aligned term. 
The strategies are denoted as AN\_weaksent in Table \ref{table:devprf1} and Table \ref{table:distribution}.

\paragraph{Interrogative to Declarative Transformation (+ Antonym/Neg).} To capture this strategy we need to determine first if the verbal irony was expressed as a rhetorical question. 
To build a classifier to detect $RQ$, we collect two categories of tweets (4K each) (1) tweets labeled with \#sarcasm or \#irony that also contain ``?'', and (2) information seeking tweets containing ``?''. We  train a binary classifier using SVM RBF Kernel with default parameters. The features are Twitter-trained word embeddings \cite{ghoshguomuresan2015}, modal verbs, pronouns, interrogative words, negations, and position of ``?'' in a tweet. We evaluate the training model on the $dev$ data and the P/R/F1 are 53.2\%, 65.4\%, and 58.6\%, respectively (in future work we plan to develop more accurate models for $RQ$ detection). Once we detect the ironic message was expressed as a $RQ$, we identify the specific interpretation strategy accompanying the transformation from interrogative to declarative form: antonym or negation. 
These combined strategies are denoted as AN$_{I\rightarrow D}$ in Table \ref{table:devprf1} and Table \ref{table:distribution}. 

\paragraph{Desiderative Constructions:} Currently, we use a simple regular  expression ``I [w]$*$ wish'' to capture counterfactual cases (AN\_{desiderative} in Tables \ref{table:devprf1} and Table  \ref{table:distribution}).


Note, when the \emph{Simple negation} and \emph{lexical antonyms} strategies are combined with other strategy (e.g., removing of intensifier), we consider this combined strategy for the interpretation of verbal irony and not the \emph{simple negation} or \emph{lexical antonym} strategy (i.e., we do not double count).

\paragraph{Phrasal antonyms and pragmatic inference:} Identifying phrasal antonyms and pragmatic inference is a complex task, and thus we propose a method of phrase matching based on phrase extraction via unsupervised alignment technique in SMT. 
We use IBM Model 4 with HMM (Giza++; \cite{och2giza}), phrase extraction via Moses \cite{koehnmoses} and the IRST tool to build the required language models. 
As post-processing, we first remove phrase pairs obtained from the S$_{im}$-H$_{int}$ bitext that are also present in the set of extracted phrases from the H$_{int}$-H$_{int}$ bitext. This increases the likelihood of retaining semantically opposite phrases, since phrases extracted from the H$_{int}$-H$_{int}$ bitext are more likely to be paraphrastic. Second, based on the translation probability scores $\phi$, for phrase \emph{e} if we have a set of aligned phrases \emph{$f_{set}$} we reject phrases that have $\phi$ scores less than $\frac{1}{size(f_{set})}$. Finally, 11,200 phrases are extracted from the S$_{im}$-H$_{int}$ bitext. The low recall for this strategy is expected since there are too many ways that users can employ pragmatic inference or rephrase the utterance without directly using any antonym or negation. In future, we will explore neural MT \cite{cho2014learning} and use external data to generate more phrases. Since we have not manually evaluated these phrase pairs, we only use this strategy after we have tried all the remaining strategies (AntPhrase+PragInf in Table \ref{table:devprf1} and Table \ref{table:distribution}).

\subsection{Results and Distribution of Linguistic Strategies} \label{subsection:distri}
The performance of the models is similar on both $test$ and $SIGN_{test}$ sets, showing consistently good performance (Table \ref{table:devprf1}; 90\% F1 for all strategies, except the AntPhrase+PragInf and AN$_{I \rightarrow D}$). Given these results, we can now apply these models to study the distribution of these strategies in the entire datasets (Table \ref{table:distribution}). The strategy distribution between our dataset S$_{im}$-H$_{int}$ and $SIGN$ dataset is similar and matches the distribution on the manual annotations on the $dev$ dataset in Table \ref{table:distrigold}.  
\begin{table}
\centering
\begin{small}
\begin{tabular}{p{2.9cm}p{1.8cm}p{1.8cm}}
\hline
Strategies & S$_{im}$-H$_{int}$ & $SIGN$  \\
\hline
Lex\_ant	& 2,198 (40.0) & 9,691 (51.8)\\
Simple\_neg &	1,596 (29.1) & 3,827 (20.5)  \\
AN\_weaksent &	895 (16.3) & 2,160 (11.6)  \\
AN$_{I \rightarrow D}$ &  329 (6.0) & 933 (5.0) \\
AN\_desiderative &	92 (1.7) & 86 (0.5)\\
AntPhrase+PragInf & 357 (6.5) & 1912 (10.1)   \\
\hline
\end{tabular}
\caption{Distribution of interpretation strategies on two datasets (in \%)}
\label{table:distribution}
\end{small}
\end{table}
The sum of the strategies can exceed the total number of the pairs since a tweet can contain several ironic sentences that are interpreted by Turkers.  For instance, in ``Dave too \textbf{nice} \dots a \textbf{nice} fella'' $\rightarrow$ ``Dave not nice \dots a mean fella'' we observe the application of two strategies, \emph{lexical antonyms} (e.g., \textbf{nice} $\rightarrow$ \textbf{mean}) and \emph{negation} (e.g., \textbf{nice} $\rightarrow$ \textbf{not nice}). 
\section{Discussion}
\subsection {Hearer-dependent Interpretation Strategies} \label{section:users}


We investigate how hearers adopt strategies for interpreting the speaker's ironic intent. To implement this study, we selected three Turkers (e.g., H$^{1}$, H$^{2}$, and H$^{3}$; In Table \ref{table:sarcesboth}, H$^{i}_{int}$ are generated by the correspondent Turker H$^{i}$), from our crowdsourced data, who were able to rephrase at least five hundred identical S$_{im}$ messages.
Note, we cannot carry this experiment on the $SIGN$ dataset \cite{peled2017sarcasm} because the annotators' information is absent there.

Although the three Turkers choose \textit{lexical antonym} and \textit{simple negation} as two top choices, there is some variation among them. 
H$^{1}$ and H$^{2}$ choose \textit{antonyms} more frequently than \textit{negation} while in contrary Turker H$^{3}$ choose \textit{negation} more than \textit{antonyms}, sometime combined with the  \textit{weakening of sentiment} strategy. As we mentioned in Section \ref{section:theory}, antonyms and direct negation are not semantically equivalent strategies since the latter, allows a graded interpretation: if ``x is not inspiring", it is not necessarily bad, but simply ``x $<$ inspiring'' \cite{giora1995irony}. 
In Table \ref{table:sarcesboth}, the S$_{im}$-H$_{int}$ pair ``passionate'' $\rightarrow$  ``boring'' and ``flattering'' $\rightarrow$ ``gross'' (interpretation of H$^{1}$) have more contrast than the pair ``passionate'' $\rightarrow$ ``not passionate'' and ``so flattering'' $\rightarrow$ ``not flattering'' (interpretation of H$^{3}$). This suggests that H$^{1}$ perceive the intensity of negative sentiment towards the target of irony (``Ed Davey'' and ``picture of dead animals'', respectively) higher than Turker H$^{3}$. 
All three Turkers have chosen the remaining strategies with similar frequencies. 
\label{section:tweets}
\subsection{Message-dependent Interpretation Strategies}  \label{section:tweets}

\paragraph{Interpretation Strategies and the Type of Semantic Incongruity:} \label{section:propstrategies} 
We investigate whether the type of semantic incongruity in the ironic message (explicit vs. implicit; see Section \ref{section:ExpIm}) influences the choice of interpretation strategies by the hearers.  To do this, we looked at S$_{im}$-level distribution of interpretation strategies used by the hearers for the same ironic message S$_{im}$.
Table \ref{table:propstrategies} represents the correlation of linguistic strategies with the type of semantic incongruity (explicit vs. implicit) as well as the presence and absence of irony markers. 

\begin{table}[h]
\centering
\begin{small}
\begin{tabular}{p{3 cm}p{.6cm}p{.6cm}p{.6cm}p{.6cm}}
\hline
Strategies & \multicolumn{2}{c} {$incongruity$} & \multicolumn{2}{c} {$marker$}  \\
& $Exp.$ & $Imp.$ & $+$ & $-$ \\
\hline
Lex\_ant  &  \emph{48.5} & 34.8 & 35.7 & \emph{42.2} \\
Simple\_neg & 24.9 & 32.3 & 28.9 & 30.0 \\
AN\_weaksent &  14.3 &  17.6 & 15.7 & 16.8 \\
AN$_{I \rightarrow D}$  & 5.9 &  6.1 & 12.3  & 3.1 \\
AN\_desiderative & 1.3 & 1.9 & 0.9 & 2.0 \\
AntPhrase+PragInf & 5.2 & 7.1 & 6.2 & 6.6 \\
\hline
\end{tabular}
\caption{Rephrasing Strategies against Incongruency and Irony Markers on S$_{im}$-H$_{int}$ dataset (in \%)}
\label{table:propstrategies}
\end{small}
\end{table}
We notice that Turkers use lexical antonyms as interpretation strategy more when the semantic incongruity is explicit than implicit (48.5\% vs. 34.8\%): the presence of explicit sentiment triggered the use of the antonym strategy. In contrary they use simple negation more when the semantic incongruity is implicit than explicit. 

We also analyze the interpretation strategies w.r.t. to the presence ($+$) or absence ($-$) of irony markers. 
We implement various morpho-syntactic as well as typographic markers (similar to \cite{ghosh2018marker}) to identify the presence of markers. We observe that $Lex\_ant$ strategy is used more in cases where the markers are absent. In S$_{im}$-H$_{int}$, markers are present twice as much in the case of implicit (21\%)  than explicit incongruity (10\%). This finding validates \cite{burgers2012verbal} who argued speakers will likely use markers to signal their ironic intent in implicit incongruity.    



\begin{figure}
\centering
 \subfloat{\frame{\includegraphics[width=2.5in,height=1.3in]
 {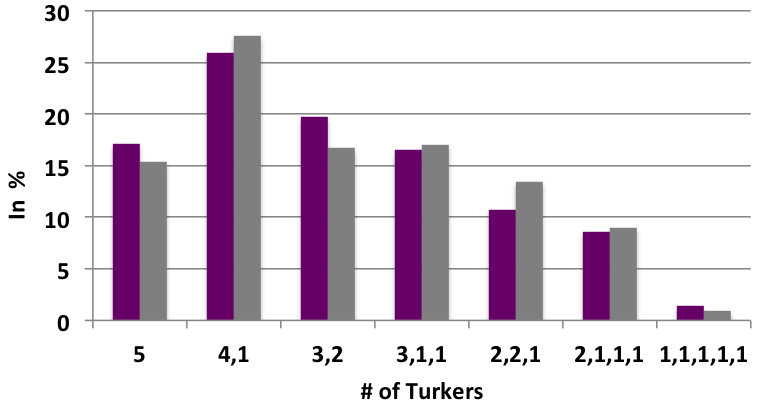}}}
 \caption{Strategies selected per message (in \%)}
\label{figure:engtweets}
\end{figure}

\paragraph{Message interpreted the same by all hearers:}
In Figure \ref{figure:engtweets}, the vertical columns (purple: S$_{im}$-H$_{int}$ and grey: $SIGN$) depict the distribution (in \%) of tweets strategy-wise. In S$_{im}$-H$_{int}$ dataset, for 17\% of messages (124 S$_{im}$s) all five Turkers use the same strategy to interpret the S$_{im}$s (labeled as \emph{5} on the X-axis), whereas for 26\% (188 S$_{im}$s), 4 Turkers used same strategy (labeled as {4,1} on X-axis) and so on.


We observe when the S$_{im}$s are marked by strong subjective words e.g., ``great'', ``best'', etc., they have been replaced in 90\% of cases as lexical antonyms (e.g., ``great'' $\rightarrow$ ``terrible''). In addition, the majority of adjectives are used in attributive position (i.e., ``\textbf{lovely} neighbor is vacuuming at night''), thus blocking paraphrases involving predicate negation. However, not all strong subjective words guarantee the use of direct opposites in the H$_{int}$s (e.g., ``flattering'' $\rightarrow$  ``not flattering''; See Table \ref{table:sarcesboth}). The choice of strategies may also depend upon the target of ironic situation  \cite{ivanko2003context}. We implement the bootstrapping algorithm from \citet{riloff} to identify ironic situations in S$_{im}$s that are rephrased by  \emph{Lexical antonym} strategy. We find utterances containing stereotypical negative situations regarding \emph{health issues} (e.g., ``having migraines'', ``getting killed by chemicals'')  
and other undesirable negative states such as ``oversleeping'', ``luggage lost'', ``stress in life'' are almost always interpreted via \emph{lexical antonym} strategy.  

Utterances where all five Turkers used \emph{simple negation}, if  negative particles are positioned in the ironic message with a sentential scope (e.g., ``not a biggie'', ``not awkward'') then they are simply omitted in the interpretations. This trend can be explained according to the inter-subjective account of negation types \cite{verhagen2005constructions}. Sentential negation leads the addressee to open up an alternative mental space where an opposite predication is at stake.

\section{Related Work}
Most NLP research on verbal irony or sarcasm has focused on the task of \emph{sarcasm detection} treating it as a binary classification task using either the utterance in isolation or adding contextual information such as conversation context, author context, visual context, or cognitive features \cite{gonzalez,liebrecht2013perfect,wallace2014humans,zhangtweet,ghosh2016fracking,schifanella2016detecting,xiong2019sarcasm,castro2019towards}. Unlike this line of work, our research focuses on how the hearer \emph{interprets} an ironic message. The findings from our study could have multiple impacts on the sarcasm detection task. First, interpretation strategies open up a scope of ``graded interpretation'' of irony instead of only a binary decision (i.e., predicting the \textbf{strength} of irony). Second, nature of semantic incongruence and stereotype irony situations can be useful features in irony detection. 

Recently, \newcite{peled2017sarcasm} proposed a computational model based on SMT to generate interpretations of sarcastic messages. We aim to deepen our understanding of such interpretations by introducing a typology of linguistic strategies. We study the distribution of these strategies via both hearer-dependent and message-dependent interpretations. 
Psycholinguistics studies that have dealt with the hearers' perception, have mainly focused on how ironic messages are processed: through the analysis of reaction times  \cite{gibbs1986psycholinguistics,katz2004saying}, the role of situational context \cite{ivanko2003context} and in tackling speaker-hearer social relations by annotating ironic texts from different genres \cite{burgers2010verbal}. However, no attention has been paid to correlations between how ironic message is expressed and how it is interpreted by the hearer, including what linguistic strategies the hearers employ. 

\section{Conclusions} \label{section:conclusion}


We leveraged a crowdsourcing task to obtain a dataset of ironic utterances paired with the hearer's verbalization of their interpretation.  We proposed a typology of linguistic strategies for verbal irony interpretation and designed computational models to capture these strategies with good performance. Our study shows (1) Turkers mostly adopt lexical antonym and negation strategies to interpret speaker's irony, (2) interpretations are correlated to stereotype ironic situations, and (3) irony expression (explicit vs. implicit incongruity and absence or presence of markers) influences the choice of interpretation strategies and match with different explanatory theories (the Gricean approach links up better with explicit incongruity, while \textit{Relevance Theory} with the implicit one). The latter can have an impact on irony detection by bringing out more discriminative semantic and pragmatic features.  

\section*{Acknowledgements}
We thank Rituparna Mukherjee, Daniel Chaparro, Pedro P\'erez S\'anchez, and Renato Augusto Vieira Nishimori who helped us in annotating as well as in running experiments. This paper partially based on the work supported by the DARPA-DEFT program. The views expressed are those of the authors and do not reflect the official policy or position of the Department of Defense or the U.S. Government. 

\bibliography{scil2020resource}
\bibliographystyle{acl_natbib}

\end{document}